%% file: zz-conferencepaper.tex
\documentclass[runningheads]{llncs}
\usepackage{textcomp}
\def\BibTeX{{\rm B\kern-.05em{\sc i\kern-.025em b}\kern-.08em
    T\kern-.1667em\lower.7ex\hbox{E}\kern-.125emX}}
\usepackage{amsmath, amsfonts}
\usepackage{epsfig}
\usepackage[mathscr]{euscript}
\usepackage{color}
\usepackage{booktabs} 
\usepackage{stmaryrd} 
\usepackage{bbm} 
\usepackage{blindtext, comment}
\usepackage{tikz}
\usepackage{standalone}
\usetikzlibrary{arrows, patterns}
\usetikzlibrary{decorations.pathreplacing,angles,quotes}
\usepackage{framed, standalone}
\usepackage{graphicx}
\usepackage{capt-of}
\usepackage{balance}
\usepackage{makecell} 
\usepackage[linesnumbered,lined,boxed,commentsnumbered,ruled,vlined]{algorithm2e}
\usepackage{setspace}
\begin{document}
\title{Predicting Sparse Clients' Actions with CPOPT-Net in the Banking Environment}
\titlerunning{Predicting Sparse Clients' Actions in the Banking Environment}
\author{\Large{\textit{Anonymous\inst{}}}}
\institute{}
\author{Jeremy Charlier\inst{1} \and Radu State\inst{1} \and Jean Hilger \inst{2}}
\authorrunning{J. Charlier et al.}
\institute{
University of Luxembourg, L-1855 Luxembourg, Luxembourg\\
\email{\{name.surname\}@uni.lu}\\
\and
BCEE, Avenue de la liberte, L-1930 Luxembourg, Luxembourg\\
\email{j.hilger@bcee.lu}}

\maketitle              
\begin{abstract}
\input{za-abstract}
\end{abstract}
\input{zb_content}
\bibliographystyle{./splncs05}
\bibliography{./zzz-mybibliography}
\end{document}

%% file: za-abstract.tex
The digital revolution of the banking system with evolving European regulations have pushed the major banking actors to innovate by a newly use of their clients' digital information. Given highly sparse client activities, we propose CPOPT-Net, an algorithm that combines the CP canonical tensor decomposition, a multidimensional matrix decomposition that factorizes a tensor as the sum of rank-one tensors, and neural networks. CPOPT-Net removes efficiently sparse information with a gradient-based resolution while relying on neural networks for time series predictions. Our experiments show that CPOPT-Net is capable to perform accurate predictions of the clients' actions in the context of personalized recommendation. CPOPT-Net is the first algorithm to use non-linear conjugate gradient tensor resolution with neural networks to propose predictions of financial activities on a public data set.

%% file: zb_content.tex
\section{Motivation}  \label{sec:section1}
\begin{figure}[t!]
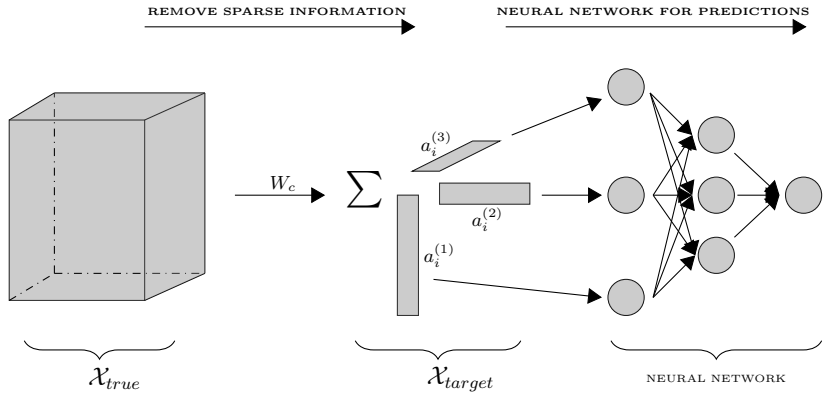

\begin{center}
\includestandalone{tikz/01-CPOPT-Net}
  \caption{In CPOPT-Net, the function $W_c$ between the original tensor $\mathcal{X}_{true}$ and the decomposed tensor $\mathcal{X}_{target}$ is minimized. Then, the latent factor vectors $\textbf{a}^{(1)}, \textbf{a}^{(2)}, \textbf{a}^{(3)}$ of each order are sent as input to the neural network. Following the neural network training, CPOPT-Net is able to predict the financial activities of the bank's clients.}
  \vspace{-0.75cm}
\label{fig::CPOPTNN}
\end{center}
\end{figure}

The modern banking environment is experiencing its own digital revolution. Strong regulatory directives are now applicable, especially in Europe with the Revised Payment Directive, PSD2, or with the General Data Protection Regulation, GDPR. Consequently, financial actors are now exploring the latest progress in data analytics and machine learning to leverage their clients' information in the context of personalized financial recommendation and client's action predictions. Recommender engines usually rely on second order matrix factorization since their accuracy has been proved in various publications \cite{brand2003fast,ghazanfar2013advantage,kumar2015role}. However, matrix factorization are limited to the unique modeling of \textit{clients} $\times$ \textit{products}. Therefore, tensor factorization have skyrocketed for the past few years \cite{lian2016regularized,zhao2016aggregated,song2017based}. Various tensor factorization, or tensor decomposition, exist for different applications \cite{kolda2009tensor,acar2011all}. However, the CP decomposition \cite{harshman1970foundations,carroll1970analysis} is the most frequently used. Two of the most popular resolution algorithms, the Alternating Least Square (ALS) \cite{harshman1970foundations,carroll1970analysis} and the non-negative ALS \cite{welling2001positive}, offer a relatively simple mathematical framework explaining its success for the new generation of recommender engines \cite{ge2016taper,almutairi2017context,cai2017heterogeneous}. In this paper, we use the gradient-based resolution for the CP decomposition \cite{acar2011scalable} to address the predictions of clients' financial activities based on time, clients' ID and transactions type. The method, illustrated in figure \ref{fig::CPOPTNN}, reduces the sparsity of the information while a neural network performs the predictions of events. We outline three contributions of our paper:

\begin{itemize}
\item We use the CP decomposition for separate modeling of each order of the data set. Since one client can have several financial activities simultaneously, we include the independent modeling of clients and financial transactions.
\item We build upon non-linear conjugate gradient resolution for the CP decomposition, CPOPT \cite{acar2011scalable}. We show CPOPT applied on a financial data set leads to small numerical errors while achieving reasonable computational time.
\item Finally, we combine CPOPT with neural network leading to CPOPT-Net. A compressed dense data set, inherited from CP, is used as an optimized input for the neural network to predict the financial activities of the clients.
\end{itemize}

The remaining of the paper is organized as follows. Section \ref{sec::propmethod} describes the CP tensor decomposition with its gradient-based resolution applied to third order financial predictions with neural network. Then, we highlight the experimental results in section \ref{sec::experiments} and we conclude by emphasizing pointers to future work.  \\

\section{CPOPT-Net and third order financial predictions} \label{sec::propmethod}
In the CP tensor decomposition \cite{harshman1970foundations,carroll1970analysis}, the tensor $\mathscr{X}\in\mathbb{R}^{I_1\times I_2\times I_3\times ...\times I_N}$ is described as the sum of the rank-one tensors
\begin{equation} \label{eq::cp}
\mathscr{X} = \sum_{r=1}^{R} \textbf{a}_r^{(1)} \circ \textbf{a}_r^{(2)} \circ \textbf{a}_r^{(3)} \circ... \circ \textbf{a}_r^{(N)}
\end{equation}
where $\textbf{a}_r^{(1)}, \textbf{a}_r^{(2)}, \textbf{a}_r^{(3)}, ..., \textbf{a}_r^{(N)}$ are vectors of size $\mathbb{R}^{I_1}, \mathbb{R}^{I_2}, \mathbb{R}^{I_3}, ..., \mathbb{R}^{I_N}$. Each vector $a_r^{(n)}$ with $n\in \left\lbrace 1, 2, ..., N \right\rbrace$ refers to one order and one rank of the tensor $\mathscr{X}$. We point out to \cite{kolda2009tensor} for further information. We use the Nonlinear Conjugate Gradient (NCG) method proposed in \cite{acar2011scalable}, CPOPT, with the strong Wolfe line search as it appears to be more stable in our case. Let $\mathcal{X}_{true}$ a real-valued \textit{N}-order tensor of size $I_1\times I _2 \times ... \times I_N$. Given $R$, the objective is to find a factorization
\begin{equation}\label{eq::xtrue_xtarget}
\mathcal{X}_{true} \approx \mathcal{X}_{target} = \sum_{r=1}^R \textbf{a}_r^{(1)} \circ ... \circ \textbf{a}_r^{(N)}
\end{equation}
with the \textit{factors} $\textbf{a}_r^{(1)}, ..., \textbf{a}_r^{(N)}$ initially randomized. Therefore, we denote by $\mathcal{X}_{target}$ the target tensor composed of the \textit{factor vectors} $\textbf{a}_r^{(1)}, ..., \textbf{a}_r^{(N)}$.  \\

The objective minimization function is denoted by $W_c(\mathcal{X}_{true}, \mathcal{X}_{target})$.
\begin{equation}\label{eq::min}
W_c(\mathcal{X}_{true}, \mathcal{X}_{target}) = \min f(\mathcal{X}_{true}, \mathcal{X}_{target}) = \dfrac{1}{2} ||\mathcal{X}_{true}-\mathcal{X}_{target}||^2
\end{equation}

The values of the factor vectors can be stacked in a parameter vector $\textbf{x}$.
\begin{equation}
\label{eq::stackx}
\textbf{x} = [\textbf{a}_1^{(1)} \cdots \textbf{a}_R^{(1)} \cdots \textbf{a}_1^{(N)} \cdots \textbf{a}_R^{(N)}]^T
\end{equation}

Therefore, we can rewrite the objective function (\ref{eq::min}) as three summands.
\begin{equation}
\label{eq::objfun}
\begin{split}
W_c(\textbf{x}) & = W_c(\mathcal{X}_{true}, \mathcal{X}_{target}) = \dfrac{1}{2} ||\mathcal{X}_{true}||^2 - \left\langle \mathcal{X}_{true}, \mathcal{X}_{target} \right\rangle + \dfrac{1}{2} ||\mathcal{X}_{target}||^2
\end{split}
\end{equation}
From (\ref{eq::objfun}), we deduce the gradient function of the CP decomposition involved in the minimization process according to the factor vectors $\textbf{a}_1^{(1)}, ..., \textbf{a}_R^{(N)}$. We refer to \cite{acar2011scalable} for more details about the gradient computation. Therefore, CPOPT-Net achieves a NCG resolution of the objective function $W_c(\mathcal{X}_{true}, \mathcal{X}_{target})$. Sparse information contained in $\mathcal{X}_{true}$ are removed in the factor vectors $\textbf{a}^{(1)}, ..., \textbf{a}^{(N)}$ of $\mathcal{X}_{target}$. Then, the factor vectors are sent as optimized inputs to the neural network. Through the training of the data set to learn the function $g(.): \mathbb{R}^3 \rightarrow \mathbb{R}^1$, the neural network is able to predict the financial activities of the bank's clients. The implementation of CPOPT-Net is summed up in algorithm \ref{algo:gradesc}.

\SetAlFnt{\scriptsize}
\SetAlCapFnt{\scriptsize}
\SetAlCapNameFnt{\scriptsize}

\begin{algorithm}[t]
\setstretch{0.85}
\DontPrintSemicolon

\KwData{tensor $\mathscr{X} \in \mathbb{R}^{I\times J\times K}$, rank R}

\KwResult{time series containing financial activities predictions, $\textbf{y}\in\mathbb{R}^1$}

\tcc{$\textbf{A} = \textbf{a}^{(1)}, \textbf{B}=\textbf{a}^{(2)}, \textbf{C}=\textbf{a}^{(3)}$}

\Begin{

random initialization \textbf{A}$\in\mathbb{R}^{I\times R}$, \textbf{B}$\in\mathbb{R}^{J\times R}$, \textbf{C}$\in\mathbb{R}^{K\times R}$

\textbf{x}$_0$ $\gets$ flatten($\textbf{A}$, $\textbf{B}$, $\textbf{C}$) as described in (\ref{eq::stackx})

$\nabla \textit{W}_{c_0} = \dfrac{\partial}{\partial x_i} W_c(\textbf{x}_0)$ $\gets$ gradient of \ref{eq::objfun} at $\textbf{x}_0$

$\alpha_0 \gets  \underset{\alpha}{\text{argmin }} f(x_0-\alpha \nabla \textit{W}_{c_0})$

$x_1 = x_0 - \alpha_0  \nabla \textit{W}_{c_0} $

$n = 0$

\Repeat{maximum number of iterations or stopping criteria}{

$\nabla \textit{W}_{c_n} = \dfrac{\partial}{\partial x_i} W_c(\textbf{x}_n)$ $\gets$ gradient of \ref{eq::objfun} at $\textbf{x}_n$

$\beta^{HS}_n \gets \dfrac{\nabla \textit{W}_{c_n}^T(-\nabla \textit{W}_{c_n}+\nabla \textit{W}_{c_{n-1}})}{s^T_{n-1}(-\nabla \textit{W}_{c_n}+\nabla \textit{W}_{c_{n-1}})}$

$s_n \gets -\nabla \textit{W}_{c_n} + \beta_n^{HS} s_{n-1}$

$\alpha_n \gets \underset{\alpha}{\text{argmin }} f(x_n-\alpha \nabla \textit{W}_{c_n})$

update $\textbf{x}_{n+1} = \textbf{x}_n - \alpha_n \nabla \textit{W}_{c_n} $

$n = n +1$

}

$\textbf{A}, \textbf{B}, \textbf{C} \gets$ unflatten($\textbf{x}_n$)

send $\textbf{A}, \textbf{B}, \textbf{C}$ to the input of the NN

training of the NN to learn the function $g(.): \mathbb{R}^3\rightarrow \mathbb{R}^1$

$\textbf{y} \in \mathbb{R}^1$ $\gets$ NN prediction of financial activities

\KwRet{$\textbf{y}\in\mathbb{R}^1$}
}

\caption{CPOPT-Net for third order financial predictions}
\label{algo:gradesc}
\end{algorithm}

\section{Predictions of clients' actions for banking recommendation} \label{sec::experiments}
\textbf{Data Availability and Experimental Setup}
In 2016, the Santander bank released an anonymized public dataset containing financial activities from its clients\footnote{The data set is available at \url{https://www.kaggle.com/c/santander-product-recommendation}}. The file contains activities of 2.5 millions of clients classified in 22 transactions labels for a 16 months period between 28 January 2015 and 28 April 2016. We choose the 200 clients having the most frequent financial activities since regular activities are more interesting for the prediction modeling. All the information is gathered in the tensor $\mathcal{X}_{true}$ of size 200$\times$22$\times$16. We define the tensor rank equal to 25. We use the Adam solver with the default parameters $\beta_1 = 0.5, \beta_2 =0.999$ for the training of the neural network\footnote{The code is available at  https://github.com/dagrate/cpoptnet.}.  \\

\textbf{Results and Discussions on CPOPT-Net}
We test CPOPT-Net using three different type of neural networks: Multi-Layer Perceptron (MLP), Convolutional Neural Network (CNN) and Long-Short Term Memory (LSTM) network. Additionally, we cross-validate the performance of the neural networks with a Decision Tree (DT). The models have been trained on one year period from 28 January 2015 until 28 January 2016. Then, the activities for the next three months are predicted with a rolling time window of one month. First, the table \ref{tab::cpoptnetals} highlight the lower numerical error obtained with the CPOPT resolution in comparison to the ALS resolution. Then, the figure \ref{fig::exp_plotpred} shows that the LSTM models the most accurately the future personal savings activities followed by the MLP, the DT, and finally the CNN. The CNN fails visually to predict accurately the savings activity in comparison to the other three methods, while the LSTM seems to achieve the most accurate predictions. We highlight this preliminary conclusion for figure \ref{fig::exp_plotpred} in table \ref{tab::predictionerrors1} by reporting four metrics: the Mean Absolute Error (MAE), the Jaccard distance, the cosine similarity and the Root Mean Square Error (RMSE). In table \ref{tab::predictionerrors2}, we show the aggregated metrics among all transaction predictions. In all the experiments, the LSTM network predicts the activities the most accurately, followed by the MLP, the DT and the CNN.

\begin{table}[t!]
 \begin{minipage}{0.325\linewidth}
  \caption{Residual errors of the objective function $W_c$ between CPOPT-Net resolution and ALS resolution at convergence (the smaller, the better). Both methods have similar computation time.}
  \centering
  \label{tab::cpoptnetals}
  \scalebox{0.775}{
  \begin{tabular}{ccc}
  \toprule
  & CPOPT-Net & CP-ALS  \\
  \midrule
 $W_c$ Error  & \textbf{10.099} & 15.896 \\
  \bottomrule
  \end{tabular} }
 \end{minipage}\hfill
 \begin{minipage}{0.625\linewidth}
  \centering
  \includegraphics[scale=0.35]{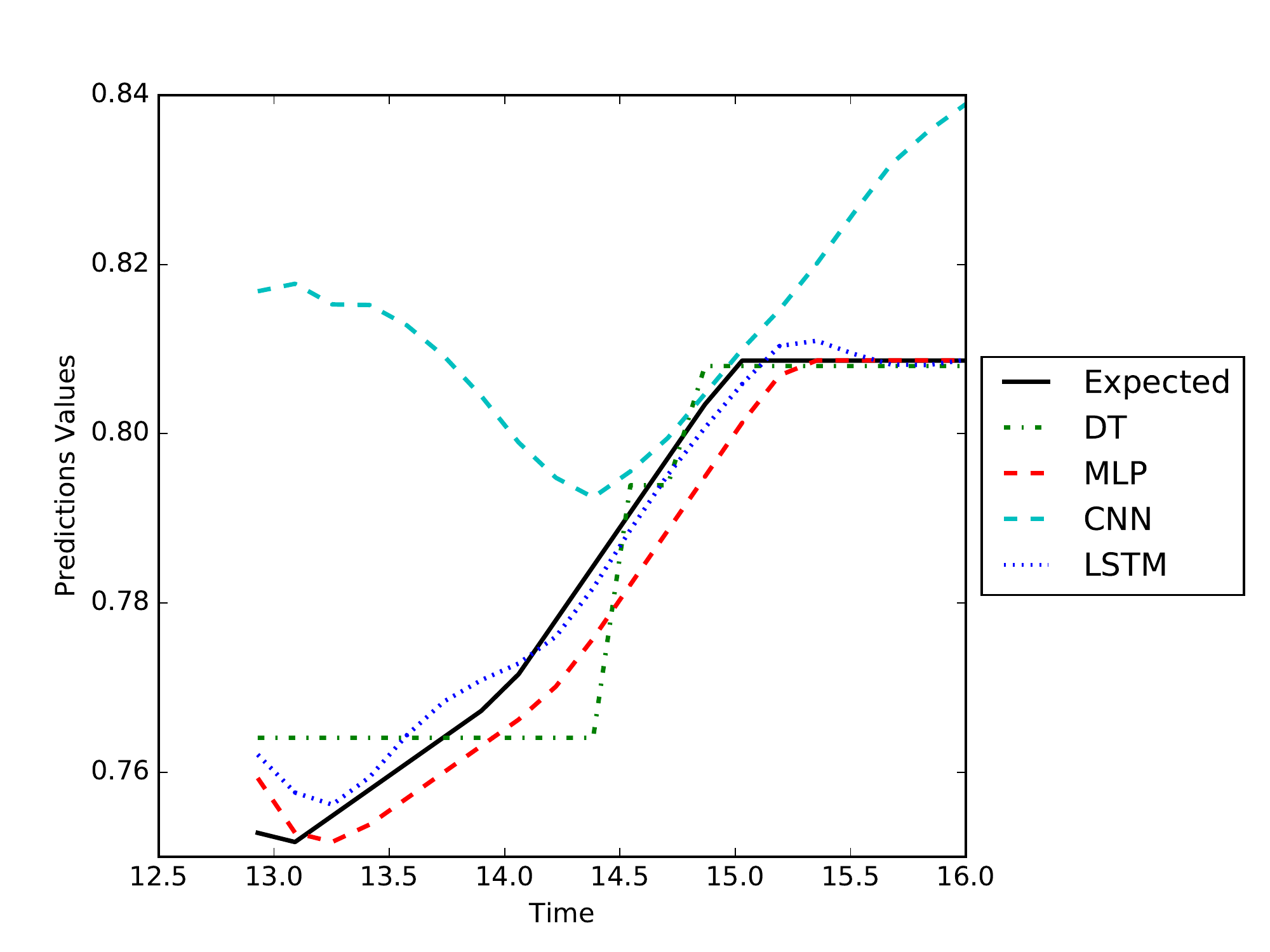}
  \captionof{figure}{Three months prediction of the evolution of the personal savings of one client. We can observe the difference of CPOPT-Net depending on the neural network chosen for the predictions.}
  \label{fig::exp_plotpred}
 \end{minipage}
\end{table}

\begin{table}[t]
\begin{minipage}{0.475\textwidth}
 \caption{Latent predictions errors on personal savings. LSTM achieves superior performance.}
 \centering
 \label{tab::predictionerrors1}
 \scalebox{0.9}{
 \begin{tabular}{ccccc}
 \toprule
 Error Measure & DT & MLP & CNN & LSTM \\
 \midrule
 MAE & 0.044 & 0.004 & 0.282 & \textbf{0.002} \\
 Jaccard dist. & 0.053 & 0.027 & 0.348 & \textbf{0.003} \\
 cosine sim. & 0.967 & 0.953 & 0.966 & \textbf{0.969}\\
 RMSE & 0.047 & 0.031 & 0.354 & \textbf{0.003} \\
 \bottomrule
 \end{tabular}}
\end{minipage} \hfill
\begin{minipage}{0.475\textwidth}
 \caption{Aggregated predictions errors on all transactions. LSTM achieves superior performance.}
 \centering
 \label{tab::predictionerrors2}
 \scalebox{0.9}{
 \begin{tabular}{ccccc}
 \toprule
 Error Measure & DT & MLP & CNN & LSTM \\
 \midrule
 MAE & 0.029 & 0.027 & 0.272 & \textbf{0.014} \\
 Jaccard dist. & 0.034 & 0.032 & 0.290 & \textbf{0.018} \\
 cosine sim. & 0.827 & 0.909 & 0.880 & \textbf{0.965}\\
 RMSE & 0.033 & 0.030 & 0.290 & \textbf{0.017} \\
 \bottomrule
 \end{tabular}}
\end{minipage}
\end{table}

\section{Conclusion}
Building upon the CP tensor decomposition, the non-linear conjugate gradient resolution and the neural networks, we propose CPOPT-Net, a predictive method for the banking industry in which the sparsity of the financial transactions is removed before performing the predictions on future clients' transactions. We conducted experiments on a public data set highlighting the prediction differences depending on the neural network involved in CPOPT-Net. Due to the recurrent activities of most of the financial transactions, we underlined the best results were found when CPOPT-Net was used with LSTM. Future work will concentrate on a limited memory resolution for a usage on very large data sets. Furthermore, the personal financial recommendation will be assessed on smaller time frame discretization, weekly or daily, with other financial transactions. It will offer a larger choice of financial product recommendations depending on the clients' mid-term and long-term interests.

%% file: tikz/01-CPOPT-Net.tex
\begin{tikzpicture}[scale=0.8, transform shape]
\draw (-7.25,-0.55) rectangle (-5,2.45);
\fill [gray, opacity=.4] (-7.25,-0.55) rectangle (-5,2.45);
\draw (-5,2.45) -- (-4,2.9);
\draw (-7.25,2.45) -- (-6.5,2.9);
\draw (-5,-0.55) -- (-4,-0.1);
\draw (-6.5,2.9) -- (-4,2.9);
\draw (-4,-0.1) -- (-4,2.9);
\draw [dashdotted] (-6.5,-0.1) -- (-4,-0.1);
\draw [dashdotted] (-6.5,-0.1) -- (-6.5,2.9);
\draw [dashdotted] (-7.25,-0.55) -- (-6.5,-0.1);

\fill [gray, opacity=.4] (-4,-0.1) -- (-4,2.9) -- (-5,2.45) -- (-5,-0.55)  -- cycle;
\fill [gray, opacity=.4] (-5,2.45) -- (-4,2.9) -- (-6.5,2.9) -- (-7.25,2.45)  -- cycle;

\coordinate (A) at (-0.8,-0.8) {} {} {} {};
\coordinate (B) at (-0.45,1.2) {} {} {} {};
\coordinate (C) at (-0.1,1.05) {} {} {} {};
\coordinate (D) at (1.4,1.4) {} {} {} {};
\coordinate (e1) at (-0.55,1.6) {} {} {} {};
\coordinate (e2) at (-0.1,1.6) {} {} {} {};
\coordinate (e3) at (0.9,2.1) {} {} {} {};
\coordinate (e4) at (0.45,2.1) {} {} {} {};
\draw (A) rectangle (B);
\fill [gray, opacity=.4] (A) rectangle (B);
\draw (C) rectangle (D);
\fill [gray, opacity=.4] (C) rectangle (D);
\draw (e1) -- (e2) -- (e3) -- (e4) -- cycle;
\fill [gray, opacity=.4] (e1) -- (e2) -- (e3) -- (e4) -- cycle;

\draw[- triangle 60] (-3.5,1.2) node (v1) {} -- (-2,1.2);

\draw  (3,3) circle (0.3);
\fill [gray, opacity=.4]  (3,3) circle (0.3);
\draw  (3,1.2) circle (0.3);
\fill [gray, opacity=.4]  (3,1.2) circle (0.3);
\draw  (3,-0.5) circle (0.3);
\fill [gray, opacity=.4]  (3,-0.5) circle (0.3);

\draw  (4.5,2.2) circle (0.3);
\fill [gray, opacity=.4]  (4.5,2.2) circle (0.3);
\draw  (4.5,1.2) circle (0.3);
\fill [gray, opacity=.4]  (4.5,1.2) circle (0.3);
\draw  (4.5,0.2) circle (0.3);
\fill [gray, opacity=.4]  (4.5,0.2) circle (0.3);

\draw  (5.95,1.2) circle (0.3);
\fill [gray, opacity=.4]  (5.95,1.2) circle (0.3);

\draw[- triangle 60] (1.1,2.2) -- (2.6,2.8);
\draw[- triangle 60] (1.6,1.2) -- (2.6,1.2);
\draw[- triangle 60] (-0.2,-0.2) -- (2.6,-0.5);

\draw[- triangle 60] (3.4,2.9) -- (4.1,2.2);
\draw[- triangle 60] (3.4,2.9) -- (4.1,1.2);
\draw[- triangle 60] (3.4,2.9) -- (4.15,0.2);

\draw[- triangle 60] (3.45,1.2) -- (4.1,2.2);
\draw[- triangle 60] (3.45,1.2) -- (4.1,1.2);
\draw[- triangle 60] (3.45,1.2) -- (4.15,0.2);

\draw[- triangle 60] (3.45,-0.5) -- (4.15,0.2);
\draw[- triangle 60] (3.45,-0.5) -- (4.1,2.2);
\draw[- triangle 60] (3.45,-0.5) -- (4.1,1.2);

\draw[- triangle 60] (4.8,1.9) -- (5.55,1.2);
\draw[- triangle 60] (4.85,1.2) -- (5.55,1.2);
\draw[- triangle 60] (4.8,0.5) -- (5.55,1.2);

\draw[- triangle 60] (-5,4) -- (-0.5,4);
\draw[- triangle 60] (1,4) -- (6,4);

\draw[decoration={brace, mirror,raise=5pt, amplitude=7pt},decorate] (-7,-1.0) -- (-4.5,-1.0);
\draw[decoration={brace, mirror,raise=5pt, amplitude=7pt},decorate] (-1.5,-1.0) -- (1.25,-1.0);
\draw[decoration={brace, mirror,raise=5pt, amplitude=7pt},decorate] (2.75,-1.0) -- (6.25,-1.0);

\node (none) at (-5.5,-1.85) {\large $\mathcal{X}_{true}$};
\node (none) at (-1.35,1.3) {\LARGE $\sum$};
\node (none) at (0.25,-1.85) {\large $\mathcal{X}_{target}$};
\node (none) at (4.5,-1.85) {\tiny NEURAL NETWORK};

\node (none) at (-0.1,0.2) {\small $a_i^{(1)}$};
\node (none) at (0.7,0.8) {\small $a_i^{(2)}$};
\node (none) at (-0.15,2.05) {\small $a_i^{(3)}$};

\node (none) at (-2.7,1.4) {\small $W_c$};

\node (none) at (3.45,4.25) {\tiny \textbf{NEURAL NETWORK FOR PREDICTIONS}};
\node (none) at (-2.8,4.25) {\tiny \textbf{REMOVE SPARSE INFORMATION}};
\end{tikzpicture}

%% file: zz-conferencepaper.bbl
\begin{thebibliography}{10}
\providecommand{\url}[1]{\texttt{#1}}
\providecommand{\urlprefix}{URL }
\providecommand{\doi}[1]{https://doi.org/#1}

\bibitem{brand2003fast}
Brand, M.: Fast online svd revisions for lightweight recommender systems. In:
  Proceedings of the 2003 SIAM International Conference on Data Mining. pp.
  37--46. SIAM (2003)

\bibitem{ghazanfar2013advantage}
Ghazanfar, M.A., Prugel, A.: The advantage of careful imputation sources in
  sparse data-environment of recommender systems: Generating improved svd-based
  recommendations. Informatica  \textbf{37}(1) (2013)

\bibitem{kumar2015role}
kumar Bokde, D., Girase, S., Mukhopadhyay, D.: Role of matrix factorization
  model in collaborative filtering algorithm: A survey. CoRR, abs/1503.07475
  (2015)

\bibitem{lian2016regularized}
Lian, D., Zhang, Z., Ge, Y., Zhang, F., Yuan, N.J., Xie, X.: Regularized
  content-aware tensor factorization meets temporal-aware location
  recommendation. In: Data Mining (ICDM), 2016 IEEE 16th International
  Conference on. pp. 1029--1034. IEEE (2016)

\bibitem{zhao2016aggregated}
Zhao, S., Lyu, M.R., King, I.: Aggregated temporal tensor factorization model
  for point-of-interest recommendation. In: International Conference on Neural
  Information Processing. pp. 450--458. Springer (2016)

\bibitem{song2017based}
Song, T., Peng, Z., Wang, S., Fu, W., Hong, X., Philip, S.Y.: Based
  cross-domain recommendation through joint tensor factorization. In:
  International Conference on Database Systems for Advanced Applications. pp.
  525--540. Springer (2017)

\bibitem{kolda2009tensor}
Kolda, T.G., Bader, B.W.: Tensor decompositions and applications. SIAM review
  \textbf{51}(3) (2009)

\bibitem{acar2011all}
Acar, E., Kolda, T.G., Dunlavy, D.M.: All-at-once optimization for coupled
  matrix and tensor factorizations. arXiv preprint arXiv:1105.3422  (2011)

\bibitem{harshman1970foundations}
Harshman, R.A.: Foundations of the parafac procedure: Models and conditions for
  an explanatory multimodal factor analysis  (1970)

\bibitem{carroll1970analysis}
Carroll, J.D., Chang, J.J.: Analysis of individual differences in
  multidimensional scaling via an n-way generalization of “eckart-young”
  decomposition. Psychometrika  \textbf{35}(3) (1970)

\bibitem{welling2001positive}
Welling, M., Weber, M.: Positive tensor factorization. Pattern Recognition
  Letters  \textbf{22}(12),  1255--1261 (2001)

\bibitem{ge2016taper}
Ge, H., Caverlee, J., Lu, H.: Taper: A contextual tensor-based approach for
  personalized expert recommendation. In: Proceedings of the 10th ACM
  Conference on Recommender Systems. pp. 261--268. ACM (2016)

\bibitem{almutairi2017context}
Almutairi, F.M., Sidiropoulos, N.D., Karypis, G.: Context-aware
  recommendation-based learning analytics using tensor and coupled matrix
  factorization. IEEE Journal of Selected Topics in Signal Processing
  \textbf{11}(5),  729--741 (2017)

\bibitem{cai2017heterogeneous}
Cai, G., Gu, W.: Heterogeneous context-aware recommendation algorithm with
  semi-supervised tensor factorization. In: International Conference on
  Intelligent Data Engineering and Automated Learning. pp. 232--241. Springer
  (2017)

\bibitem{acar2011scalable}
Acar, E., Dunlavy, D.M., Kolda, T.G.: A scalable optimization approach for
  fitting canonical tensor decompositions. Journal of Chemometrics
  \textbf{25}(2) (2011)

\end{thebibliography}
